\documentclass[11pt,a4paper]{article}
\usepackage[hyperref]{emnlp-ijcnlp-2019}
\usepackage{times}
\usepackage{latexsym}

\usepackage{url}

\aclfinalcopy % Uncomment this line for the final submission

\usepackage{amsmath}
\usepackage{cleveref}

\usepackage{amsfonts}
\usepackage{xcolor}

\newcommand{\keyword}[1]{\textit{#1}}

\usepackage{tabu}
\usepackage{booktabs}
\usepackage{multirow}
\usepackage{subcaption}

\usepackage{mathtools}

\DeclarePairedDelimiter{\norm}{\lvert\lvert}{\rvert\rvert}

\usepackage{bm}
\def\eqref#1{equation~\ref{#1}}
\def\1{\bm{1}}

\def\vw{{\bm{w}}}
\def\vx{{\bm{x}}}

\def\mW{{\bm{W}}}

\DeclareMathAlphabet{\mathsfit}{\encodingdefault}{\sfdefault}{m}{sl}
\SetMathAlphabet{\mathsfit}{bold}{\encodingdefault}{\sfdefault}{bx}{n}

\def\sR{{\mathbb{R}}}

\usepackage[warn]{textcomp}
\usepackage{listings}
\newcommand{\arTOen}{ar\textrightarrow en}
\newcommand{\enTOhe}{en\textrightarrow he}
\newcommand{\enTOvi}{\textit{en\textrightarrow vi}}
\newcommand{\glTOen}{gl\textrightarrow en}
\newcommand{\skTOen}{sk\textrightarrow en}

\newcommand{\LNorm}{\textsc{LayerNorm}}
\newcommand{\SCNorm}{\textsc{ScaleNorm}}
\newcommand{\SCNormOne}{\textsc{ScaleNorm} ($g$=$\sqrt{d}$)}
\newcommand{\RMSNorm}{\textsc{RMSNorm}}
\newcommand{\PreNorm}{\textsc{PreNorm}}
\newcommand{\PostNorm}{\textsc{PostNorm}}
\newcommand{\FairSeq}{\textsc{fairseq}}
\newcommand{\fixnorm}{\textsc{FixNorm}}
\newcommand{\radam}{\textsc{RAdam}}
\newcommand{\fixup}{\textsc{Fixup}}

\newcommand{\ORGLR}{\textsc{InvSqrtDecay}}
\newcommand{\NOWARMUP}{\textsc{NoWarmup}}
\newcommand{\VALBASED}{\textsc{ValDecay}}

\newcommand{\WMT}{WMT\,\textquotesingle 14}
\newcommand{\IWSLT}{IWSLT\,\textquotesingle 15}

\newcommand{\SmallInit}{\textsc{SmallInit}}
\newcommand{\lrtwo}{\textsc{2$\times$LR}}

\renewcommand{\paragraph}[1]{\par\medskip\noindent\textbf{#1}}

\newcommand*\samethanks[1][\value{footnote}]{\hyperlink{*}{\footnotemark[#1]}}

\title{Transformers without Tears:\\Improving the Normalization of Self-Attention}

\author{Toan Q. Nguyen\Thanks{\ Equal contribution.} \\
  University of Notre Dame\Thanks{\ Work done during an internship at Amazon AWS AI.} \\
  {\tt tnguye28@nd.edu} \\\And
  Julian Salazar\samethanks[1]\\
  Amazon AWS AI \\
  {\tt julsal@amazon.com} \\
  }

\date{}

\begin{document}
\maketitle

\begin{abstract}
We evaluate three simple, normalization-centric changes to improve  Transformer training. First, we show that pre-norm residual connections (\PreNorm) and smaller initializations enable warmup-free, validation-based training with large learning rates. Second, we propose $\ell_2$ normalization with a single scale parameter (\SCNorm) for faster training and better performance. Finally, we reaffirm the effectiveness of normalizing word embeddings to a fixed length (\fixnorm). On five low-resource translation pairs from TED Talks-based corpora, these changes always converge, giving an average +1.1 BLEU over state-of-the-art bilingual baselines and a new 32.8 BLEU on \IWSLT\ English-Vietnamese. We observe sharper performance curves, more consistent gradient norms, and a linear relationship between activation scaling and decoder depth. Surprisingly, in the high-resource setting (\WMT\ English-German), \SCNorm\ and \fixnorm\ remain competitive but \PreNorm\ degrades performance. Preprocessing scripts and code are released at\\ \url{https://github.com/tnq177/transformers_without_tears}.
\end{abstract}

\section{Introduction}
The Transformer \cite{NIPS2017_7181} has become the dominant architecture for neural machine translation (NMT) due to its train-time parallelism and strong downstream performance. Various modifications have been proposed to improve the efficiency of its multi-head attention and feedforward sublayers \cite{Guo2019,Sukhbaatar2019}. Our work focuses on \keyword{layer normalization} (\LNorm) \cite{Ba2015}, which we show has an outsized role in the convergence and performance of the Transformer in two ways:

\paragraph{Placement of normalization.} The original Transformer uses \keyword{post-norm residual units} (\PostNorm), where layer normalization occurs after the sublayer and residual addition. However, \citet{Chen2018} found that \keyword{pre-norm residual units} (\PreNorm), where layer normalization occurs immediately before the sublayer, were instrumental to their model's performance. \citet{Wang2019-learning-deep-transformers} compare the two, showing that \PreNorm\ makes backpropagation more efficient over depth and training Transformers with deep, 30-layer encoders.

Our work demonstrates additional consequences in the base ($\le$6-layer encoder) Transformer regime. We show that \PreNorm\ enables warmup-free, validation-based training with large learning rates even for small batches, in contrast to past work on scaling NMT \cite{Ott2018}. We also partly reclaim \PostNorm's stability via smaller initializations, although \PreNorm\ is less sensitive to this magnitude and can improve performance. However, despite \PreNorm's recent adoption in many NMT frameworks, we find it degrades base Transformer performance on \WMT\ English-German.

\paragraph{Choice of normalization.} \citet{Santurkar2018} show that batch normalization's effectiveness is not from reducing internal covariate shift, but from smoothing the loss landscape. They achieve similar or better performance with non-variance-based normalizations in image classification. Hence, we propose replacing \LNorm\ with the simpler \keyword{scaled $\ell_2$ normalization} (\SCNorm), which normalizes activation vectors to a \keyword{single} learned length $g$. This is both inspired by and synergistic with jointly fixing the word embedding lengths (\fixnorm) \cite{Nguyen2018-improving-lexical-choice}. These changes improve the training speed and low-resource performance of the Transformer without affecting high-resource performance.
\par\medskip

On five low-resource pairs from the TED Talks \cite{Qi2018-word-embeddings-nmt} and \IWSLT\ \cite{Cettolo2015} corpora, we first train state-of-the-art Transformer models (+4.0 BLEU on average over the best published NMT bitext-only numbers). We then apply \PreNorm, \fixnorm, and \SCNorm\ for an average total improvement of +1.1 BLEU, where each addition contributes at least +0.3 BLEU (\Cref{sec:experiments}), and attain a new 32.8 BLEU on \IWSLT\ English-Vietnamese. We validate our intuitions in \Cref{sec:analysis} by showing sharper performance curves (i.e., improvements occur at earlier epochs) and more consistent gradient norms. We also examine the per-sublayer $g$'s learned by \SCNorm, which suggest future study.

\section{Background}

\subsection{Identity mappings for transformers}
\label{ssec:identity-mappings}
\keyword{Residual connections} \cite{He2016} were first introduced to facilitate the training of deep convolutional networks, where the output of the $\ell$-th layer $F_{\ell}$ is summed with its input:
\begin{equation}
    \vx_{\ell+1} = \vx_\ell + F_{\ell}(\vx_\ell).
\end{equation}
The identity term $\vx_\ell$ is crucial to greatly extending the depth of such networks \cite{He2016-identity-mappings}. If one were to scale $\vx_\ell$ by a scalar $\lambda_\ell$, then the contribution of $\vx_\ell$ to the final layer $F_L$ is $(\prod_{i=\ell}^{L-1}\lambda_i) \vx_\ell$. For deep networks with dozens or even hundreds of layers $L$, the term $\prod_{i=\ell}^{L-1}\lambda_i$ becomes very large if $\lambda_i > 1$ or very small if $\lambda_i < 1$, for enough $i$. When backpropagating from the last layer $L$ back to $\ell$, these multiplicative terms can cause exploding or vanishing gradients, respectively. Therefore they fix $\lambda_i = 1$, keeping the total residual path an identity map.

The original Transformer applies \LNorm\ after the sublayer and residual addition (\PostNorm):  
\begin{equation}\label{post-norm}
    \vx_{\ell+1} = \LNorm(\vx_\ell + F_{\ell}(\vx_\ell)).
\end{equation}
We conjecture this has caused past convergence failures \cite{Popel2018, Shazeer2018}, with {\LNorm}s in the residual path acting similarly to $\lambda_i \ne 1$; furthermore, warmup was needed to let \LNorm\ safely adjust scale during early parts of training. Inspired by \citet{He2016-identity-mappings}, we apply \LNorm\ immediately before each sublayer (\PreNorm): 
\begin{equation}\label{pre-norm}
    \vx_{\ell+1} = \vx_\ell + F_{\ell}(\LNorm(\vx_\ell)).
\end{equation}  
This is cited as a stabilizer for Transformer training \cite{Chen2018, Wang2019-learning-deep-transformers} and is already implemented in popular toolkits \cite{tensor2tensor, fairseq, sockeye}, though not necessarily used by their default recipes. \citet{Wang2019-learning-deep-transformers} make a similar argument to motivate the success of \PreNorm\ in training very deep Transformers. Note that one must append an additional normalization after both encoder and decoder so their outputs are appropriately scaled. We compare \PostNorm\ and \PreNorm\ throughout \Cref{sec:experiments}.

\subsection{Weight initialization}
\label{ssec:weight-init}

Xavier normal initialization \cite{Glorot2010} initializes a layer's weights $\mW_{\ell} \in \sR^{d_{\ell+1} \times d_{\ell}}$ ($d_{\ell}$ is the hidden dimension) with samples from a centered normal distribution with layer-dependent variance:
\begin{equation}\label{xavier}
    (\mW_{\ell})_{i,j} \sim \mathcal{N}\left(0, \sqrt{\frac{2}{d_{\ell} + d_{\ell+1}}}\right).
\end{equation}
Our experiments with this default initializer find that \PostNorm\ sometimes fails to converge, especially in our low-resource setting, even with a large number of warmup steps. One explanation is that Xavier normal yields initial weights that are too large. In implementations of the Transformer, one scales the word embeddings by a large value (e.g., $\sqrt{d} \approx 22.6$ for $d=512$), giving vectors with an expected square norm of $d$. \LNorm's unit scale at initialization preserves this same effect. Since feedforward layers already have their weights initialized to a smaller standard deviation, i.e., $\sqrt{\frac{2}{d + 4d}}$, we propose reducing the attention layers' initializations from $\sqrt{\frac{2}{d + d}}$ to $\sqrt{\frac{2}{d + 4d}}$ as well (\SmallInit), as a corresponding mitigation. We evaluate the effect of this on \PostNorm\ vs.\ \PreNorm\ in \Cref{experiment_weight_init}.

\subsection{Scaled $\ell_2$ normalization and \fixnorm}
\label{ssec:scaled-cosine}

\LNorm\ is inspired by batch normalization \cite{Ioffe2015}, both of which aim to reduce internal covariate shift by fixing the mean and variance of activation distributions. Both have been applied to self-attention \cite{NIPS2017_7181,Kool2019}. However, \citet{Santurkar2018} show that batch normalization's success has little to do with covariate shift, but comes instead from smoothing the loss landscape. For example, they divide by the pre-centered $\ell_p$ norm instead of the variance and achieve similar or better results in image classification.

Hence, we propose replacing \LNorm\ with \keyword{scaled $\ell_2$ normalization}:
\begin{equation} \label{scnorm}
    \SCNorm(\vx; g) = g\frac{\vx}{\norm{\vx}}.
\end{equation}
This can be viewed as projecting $d$-dimensional vectors onto a $(d-1)$-dimensional hypersphere with learned radius $g$. This expresses the inductive bias that each sublayer's activations has an ideal ``global scale,'' a notion we empirically validate in \Cref{ssec:g-values}. \SCNorm\ replaces the $2d$ scale and shift parameters of \LNorm\ with a single learned scalar, improving computational and parameter efficiency while potentially regularizing the loss landscape.

This bias has an explicit interpretation at the final layer: large inner products sharpen the output distribution, causing frequent words to disproportionately dominate rare words. This led \citet{Nguyen2018-improving-lexical-choice} to introduce $\fixnorm(\vw) = g \frac{\vw}{\norm{\vw}}$ with fixed $g$ at the last linear layer, to maximize the angular difference of output representations and aid rare word translation. By making $g$ learnable, we can apply \SCNorm\ and \fixnorm\ jointly, which means applying the following at the final linear layer:
\begin{equation}
\begin{split}
	(\SCNorm + &\fixnorm)(\vx, \vw; g)\\
	 &= g\frac{\vw \cdot \vx}{\norm{\vw}\norm{\vx}}.
\end{split}
\end{equation}
Note that this combination at the last layer is equivalent to cosine normalization \cite{Luo2017} with a learned scale.

%%% <MOVE AROUND FOR OPTIMAL FLOW>
\begin{table*}[ht!]
\small
\begin{minipage}{1.0\linewidth}
	\centering
	
	\begin{tabu}{@{}c|cccccccc@{}}
\toprule
      & \# egs. & \# src.\ + tgt.\ toks. & \# iters./epoch & max.\ epoch & \# enc./dec.\ layers & \# heads/layer & dropout & \# BPE \\ \midrule
\textbf{\glTOen} & 10k         & 0.37M                     & 100                 & 1000          & 4         & 4        & 0.4     & 3k     \\
\textbf{\skTOen} & 61k         & 2.32M                     & 600                 & 200           & 6         & 8        & 0.3     & 8k     \\
\textbf{\enTOvi} & 133k        & 5.99M                     & 1500                & 200           & 6         & 8        & 0.3     & 8k     \\
\textbf{\enTOhe} & 212k        & 7.88M                     & 2000                & 200           & 6         & 8        & 0.3     & 8k     \\
\textbf{\arTOen} & 214k        & 8.09M                     & 2000                & 200           & 6         & 8        & 0.3     & 8k     \\
\bottomrule
\end{tabu}
	
\end{minipage}
\caption{Data and model properties for low-resource NMT. \enTOvi\ is from IWSLT 2015; the rest are from the TED Talks corpus.}
\label{tab:stats}
\end{table*}

%%% </MOVE AROUND FOR OPTIMAL FLOW>

\subsection{Learning rates}
\label{ssec:learning-rate}

Despite using an adaptive optimizer, Adam \cite{Kingma2014}, Transformer training uses a learning rate (LR) schedule with a linear \keyword{warmup} and an inverse square root \keyword{decay} (\ORGLR):
\begin{equation} \label{xmer-lr}
    \text{LR}(n) = \frac{\lambda}{\sqrt{d}} \min\left(\frac{1}{\sqrt{n}}, \frac{n}{n_{\text{warmup}}^{1.5}}\right),
\end{equation}
where $d$ is the hidden dimension of the self-attention layers, and $\lambda$, $n_{\text{warmup}}$ are hyperparameters that determine the highest learning rate achieved and the number of steps to reach it, respectively. These two hyperparameters have been the subject of much empirical study \cite{Popel2018, Ott2018}. In light of our modifications however, we revisit various aspects of this schedule:

\paragraph{Warmup-free training.} We conjectured that warmup is primarily needed when using \PostNorm\ to gradually learn \LNorm\ parameters without gradient explosion/vanishing (\Cref{ssec:identity-mappings}). Hence, we evaluate both \PreNorm\ and \PostNorm\ without warmup in \Cref{experiment_lr}.

\paragraph{Large learning rates.} To speed up training, one often explores using larger learning rates. In the context of Transformer, \citet{Ott2018} and \citet{Aharoni2019} take $\lambda \in \{2, 3\}$ instead of the conventional $\lambda = 1$. \citet{Ott2018} showed that one can scale up Adam's learning rate to $10^{-3}$ with an extremely large batch (400k tokens). However, the improved convergence provided by our modifications could enable higher learning rates with much small batch sizes (4k tokens), as examined in \Cref{experiment_lr}.

\paragraph{Validation-based decay.} For similar reasons, one might wish to adopt a classic validation-based decay, i.e., training at a high learning rate for as long as tenable, decaying rapidly when development scores flatline. This has inspired usage of fixed decay schemes upon convergence with \ORGLR\ \cite{Dong2018, Salazar2019}. We revisit \VALBASED\ under our modifications, where we still perform a linear warmup but then multiply by a scale $\alpha_{\text{decay}} < 1$ when performance on a development set does not improve over $patience$ evaluations.

\section{Experiments and results}
\label{sec:experiments}
We train Transformer models for a diverse set of five low-resource translation pairs from the TED Talks \cite{Qi2018-word-embeddings-nmt} and the \IWSLT\ \cite{Cettolo2015} corpora. Details are summarized in \Cref{tab:stats}. For more information motivating our choice of pairs and for exact training details, refer to \Cref{appendix:setup}.

\subsection{Large vs. small initialization} \label{experiment_weight_init}
To see the impact of weight initialization, we run training on the \enTOvi\ dataset using warmup steps of {4k, 8k, 16k} (\Cref{tab:big_small_init}). With default initialization, \PostNorm\ fails to converge on this dataset even with a long warmup of 16k steps, only reaching 5.76 BLEU.

%%% <MOVE AROUND FOR OPTIMAL FLOW>
\begin{table}[h]
\small
\begin{minipage}{1.0\linewidth}
	\centering
\begin{tabu}{@{}l|l|ccc@{}}
\toprule
      \multirow{2}{*}{Xavier normal} & & \multicolumn{3}{c}{\# warmup steps}       \\
      & & 4k & 8k & 16k \\ \midrule
\multirow{2}{*}{Baseline}   & \PostNorm    & fail  & fail  & 5.76  \\
                            & \PreNorm   & 28.52 & 28.73 & 28.32 \\ \midrule
\multirow{2}{*}{\SmallInit} & \PostNorm    & 28.17 & 28.20  & 28.62 \\
                            & \PreNorm    & 28.26 & 28.44 & 28.33 \\
\bottomrule
\end{tabu}
\end{minipage}
\caption{Development BLEU on \enTOvi\ using Xavier normal initialization (baseline versus \SmallInit).}
\label{tab:big_small_init}
\end{table}
%%% </MOVE AROUND FOR OPTIMAL FLOW>

%%% <MOVE AROUND FOR OPTIMAL FLOW>
\begin{table*}[ht]
\small
\begin{minipage}{1.0\linewidth}
    \centering
\begin{tabu}{@{}r|c|c|c|c|c|c@{}}
\toprule
                 & \multicolumn{1}{c|}{\textbf{\glTOen}}       & \multicolumn{1}{c|}{\textbf{\skTOen}}       & \multicolumn{1}{c|}{\textbf{\enTOvi}}       & \multicolumn{1}{c|}{\textbf{\enTOhe}}       & \multicolumn{1}{c|}{\textbf{\arTOen}}       &
                     \multicolumn{1}{c}{\textbf{average $\Delta$}} \\ \midrule
\PostNorm\ + \LNorm\ (published) & 16.2 & 24.0 & 29.09 & 23.66 & 27.84 & -4.05 \\\midrule
\PostNorm\ + \LNorm\ (1)  & 18.47 & 29.37 & 31.94 & 27.85 & 33.39 & +0.00 \\
\PreNorm\ + \LNorm\ (2)  & 19.09 & 29.45 & 31.92 & 28.13 & 33.79 & +0.27 \\
\PreNorm\ + \fixnorm\ + \LNorm\ (3) & 19.38 & 29.50 & 32.45 & 28.39 & 34.35$^{\dagger}$ & +0.61 \\
\PreNorm\ + \fixnorm\ + \SCNorm\ (4) & 20.91$^{\ddagger \ast}$ & 30.25$^{\ddagger \ast}$ & 32.79$^{\ast}$ & 28.44$^{\ast}$ & 34.15$^{\ast}$ & +1.10 \\
\bottomrule
\end{tabu}
\end{minipage}
\caption{Test BLEU using \PostNorm\ or \PreNorm\ and different normalization techniques. Published values are from \citet{Wang2018-switchout, Neubig2019, Aharoni2019}. $\dagger$, $\ddagger$ and $\ast$ indicate significant improvement of (3) over (2), (4) over (3), and (4) over (1), respectively; $p < 0.01$ via bootstrap resampling \cite{Koehn2004}.}
\label{tab:lnorm-scnorm}
\end{table*}
%%% </MOVE AROUND FOR OPTIMAL FLOW>
%%% <MOVE AROUND FOR OPTIMAL FLOW>
\begin{table*}[ht!]
\small
\begin{minipage}{1.0\linewidth}
    \centering
\begin{tabu}{@{}l|c|c|c|c|c@{}}
\toprule
                 & \multicolumn{1}{c|}{\textbf{\glTOen}}       & \multicolumn{1}{c|}{\textbf{\skTOen}}       & \multicolumn{1}{c|}{\textbf{\enTOvi}}       & \multicolumn{1}{c|}{\textbf{\enTOhe}}       & \multicolumn{1}{c}{\textbf{\arTOen}}       \\ \midrule
\NOWARMUP\          & 18.00 & 28.92 & 28.91 & 30.33 & 35.40 \\
\ORGLR\          & 22.18 & 29.08 & 28.84 & 30.30 & 35.33 \\
\VALBASED\          & 21.45 & 29.46 & 28.67 & 30.69 & 35.46 \\
\ORGLR\ + \lrtwo & 21.92 & 29.03 & 28.76 & 30.50 & 35.33 \\
\VALBASED\ + \lrtwo & 21.63 & 29.49 & 28.46 & 30.13 & 34.95 \\
\bottomrule
\end{tabu}
\end{minipage}
\caption{Development BLEU for \PreNorm\ + \fixnorm\ + \SCNorm, trained with different learning rate schedulers.}
\label{tab:learning-rate}
\end{table*}
%%% </MOVE AROUND FOR OPTIMAL FLOW>

The second row shows that taking a smaller standard deviation on the attention weights (\SmallInit) restores convergence to \PostNorm. Though the $\sqrt{2/5} \approx 0.63$ adjustment used here seems marginal, operations like residual connections and the products between queries and keys can compound differences in scale. Though both models now achieve similar performance, we note that \PreNorm\ works in all setups, suggesting greater stability during training. For all remaining experiments, we use \PostNorm\ and \PreNorm\ with \SmallInit. We find this choice does not affect the performance of \PreNorm.

\subsection{Scaled $\ell_2$ normalization and \fixnorm}
\label{sec:experiments_scnorm}

To compare \SCNorm\ and \LNorm, we take 8k warmup steps for all further experiments. Since we tie the target input word embedding and the last linear layer's weight (\Cref{appendix:setup}), \fixnorm\ is implemented by applying $\ell_2$ normalization to the word embedding, with each component initialized uniformly in $[-0.01, 0.01]$. For non-\fixnorm\ models, word embeddings are initialized with mean 0 and standard deviation $\sqrt{1/d}$ so they sum to unit variance. All $g$'s in \SCNorm\ are initialized to $\sqrt{d}$.

\Cref{tab:lnorm-scnorm} shows our results along with some published baselines. First, note that our Transformer baselines with \PostNorm\ + \LNorm\ (1) are very strong non-multilingual NMT models on these pairs. They outperform the best published numbers, which are all Transformer models in the past year, by an average margin of +4.0 BLEU. Then, we see that \PreNorm\ (2) achieves comparable or slightly better results than \PostNorm\ on all tasks. \fixnorm\ (3) gives an additional gain, especially on \arTOen\ ($p < 0.01$).

Finally, we replace \LNorm\ with \SCNorm\ (4). \SCNorm\ significantly improves on \LNorm\ for two very low-resource pairs, \glTOen\ and \skTOen. On the other tasks, it performs comparably to \LNorm. Upon aggregating all changes, our final model with \SCNorm\ and \fixnorm\ improves over our strong baseline with \PostNorm\ on all tasks by an average of +1.1 BLEU ($p < 0.01$), with each change contributing an average of at least +0.3 BLEU. In \Cref{ssec:g-values} and \Cref{ssec:generalization}, we further examine where the performance gains of \SCNorm\ come from.

Moreover, \SCNorm\ is also faster than \LNorm. Recall that for each vector of size $d$, \LNorm\ needs to compute mean, standard deviation, scaling, and shifting, which costs $O(7d)$ operations. For \SCNorm, we only need $O(3d)$ operations to perform normalization and global scaling. This does not account for further gains due to reduction in parameters. In our implementation, training with \SCNorm\ is around 5\% faster than with \LNorm, similar to the speedups on NMT observed by \citet{Zhang2019}'s \RMSNorm\ (which can be viewed as \SCNorm\ with per-unit scales; see \Cref{ssec:g-values}).

\subsection{Learning rates} \label{experiment_lr}

We compare the original learning rate schedule in equation \ref{xmer-lr} (\ORGLR) with validation-based decay (\VALBASED), possibly with no warmup (\NOWARMUP). We use $\lambda=1$, $n_{warmup}=8\text{k}$ for \ORGLR\ and \VALBASED. For \NOWARMUP, we instead use a learning rate of $3 \cdot 10^{-4}$ for all datasets. For both \VALBASED\ and \NOWARMUP, we take $\alpha_{decay}=0.8$ and $patience=3$. For experiments with high learning rate, we use either \VALBASED\ or \ORGLR\ with $\lambda = 2$ (giving a peak learning rate of $\approx10^{-3}$). All experiments use \PreNorm\ + \fixnorm\ + \SCNorm.

In \Cref{tab:learning-rate}, we see that \NOWARMUP\ performs comparably to \ORGLR\ and \VALBASED\ except on \glTOen. We believe that in general, one can do without warmup, though it remains useful in the lowest resource settings. In our \lrtwo\ experiments, we can still attain a maximum learning rate of $10^{-3}$ without disproportionately overfitting to small datasets like \glTOen.
 
One might hypothesize that \VALBASED\ converges more quickly to better minima than \ORGLR\ by staying at high learning rates for longer. However, both schedulers achieve similar results with or without doubling the learning rate. This may be due to the tail-end behavior of \VALBASED\ methods, which can involve multiplicative decays in rapid succession. Finally, our \lrtwo\ experiments, while not yielding better performance, show that \PreNorm\ allows us to train the Transformer with a very high learning rate despite small batches (4k tokens).  

Since \PreNorm\ can train without warmup, we wonder if \PostNorm\ can do the same. We run experiments on \enTOvi\ with \NOWARMUP, varying the number of encoder/decoder layers.  As seen in \Cref{tab:no-warm-post-prev}, \PostNorm\ often fails without warmup even with 5 or 6 layers. Even at 4 layers, one achieves a subpar result compared to \PreNorm. This reaffirms \Cref{experiment_weight_init} in showing that \PreNorm\ is more stable than \PostNorm\ under different settings.
%%% <MOVE AROUND FOR OPTIMAL FLOW>
\begin{table}[h]
\small
\begin{minipage}{1.0\linewidth}
	\centering
\begin{tabu}{@{}r|ccc@{}}
\toprule
    & 4 layers & 5 layers & 6 layers              \\ \midrule
\PostNorm & 18.31 & fails & fails \\
\PreNorm & 28.33 & 28.13 & 28.32           \\
\bottomrule
\end{tabu}
\end{minipage}
\caption{Development BLEU on \enTOvi\ using \NOWARMUP, as number of encoder/decoder layers increases.}
\label{tab:no-warm-post-prev}
\end{table}
%%% </MOVE AROUND FOR OPTIMAL FLOW>  

\subsection{High-resource setting} \label{experiment_highres}

Since all preceding experiments were in low-resource settings, we examine if our claims hold in a high-resource setting. We train the Transformer base model on \WMT\ English-German using \FairSeq\ and report tokenized BLEU scores on \keyword{newstest2014}. Implementation of our methods in \FairSeq\ can be found in \Cref{listing-fairseq}.

In \Cref{tab:high-resource}, \SCNorm\ and \fixnorm\ achieve equal or better results than \LNorm. Since \SCNorm\ is also faster, we recommend using both as drop-in replacements for \LNorm\ in all settings. Surprisingly, in this task \PostNorm\ works notably better than \PreNorm; one observes similar behavior in \citet{Wang2019-learning-deep-transformers}. We speculate this is related to identity residual networks acting like shallow ensembles \cite{resnetensemble} and thus undermining the learning of the longest path; further study is required.

%%% <MOVE AROUND FOR OPTIMAL FLOW>
\begin{table}[h]
\small
\begin{minipage}{1.0\linewidth}
	\centering
\begin{tabu}{@{}r|c@{}}
\toprule
    & \multicolumn{1}{c}{\textbf{newstest2014}}       \\\midrule
\PostNorm\ + \LNorm\ (published) & 27.3 \\\midrule
\PreNorm\ + \LNorm & 26.83 \\
\PreNorm\ + \fixnorm\ + \SCNorm & 27.07 \\
\PostNorm\ + \LNorm & 27.58 \\
\PostNorm\ + \fixnorm\ + \SCNorm & 27.57 \\
\bottomrule
\end{tabu}
\end{minipage}
\caption{BLEU scores from \WMT\ English-to-German. Published value is from \citet{NIPS2017_7181}.}
\label{tab:high-resource}
\end{table}
%%% </MOVE AROUND FOR OPTIMAL FLOW>

\section{Analysis}
\label{sec:analysis}

\subsection{Performance curves}
\label{sec:perf-curve}

\begin{figure}[h!]
\centering
\includegraphics[width=0.45\textwidth]{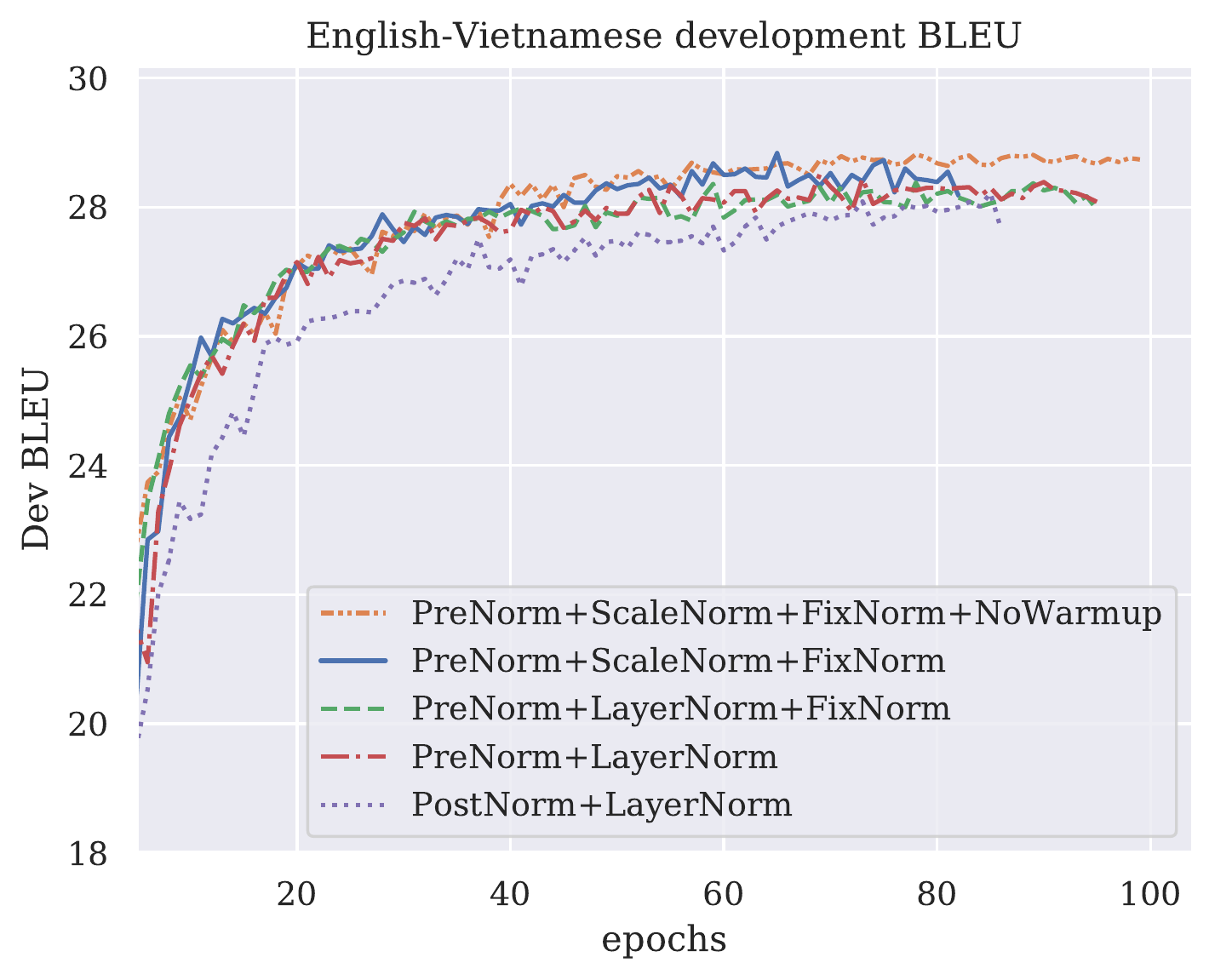}
\caption{Development BLEU on \enTOvi\ with \PostNorm\ or \PreNorm, and with \LNorm\ or \SCNorm.}
\label{fig:dev_bleus}
\end{figure}

\Cref{fig:dev_bleus} shows that \PreNorm\ not only learns faster than \PostNorm, but also outperforms it throughout training. Adding \fixnorm\ also gives faster learning at first, but only achieves close performance to that with \PreNorm\ and no \fixnorm. However, once paired with \SCNorm, we attain a better BLEU score at the end. Because of the slow warmup period, \SCNorm\ with warmup learns slower than \SCNorm\ without warmup initially; however, they all converge at about the same rate.  

\begin{figure}[h!]
\centering
\includegraphics[width=0.45\textwidth]{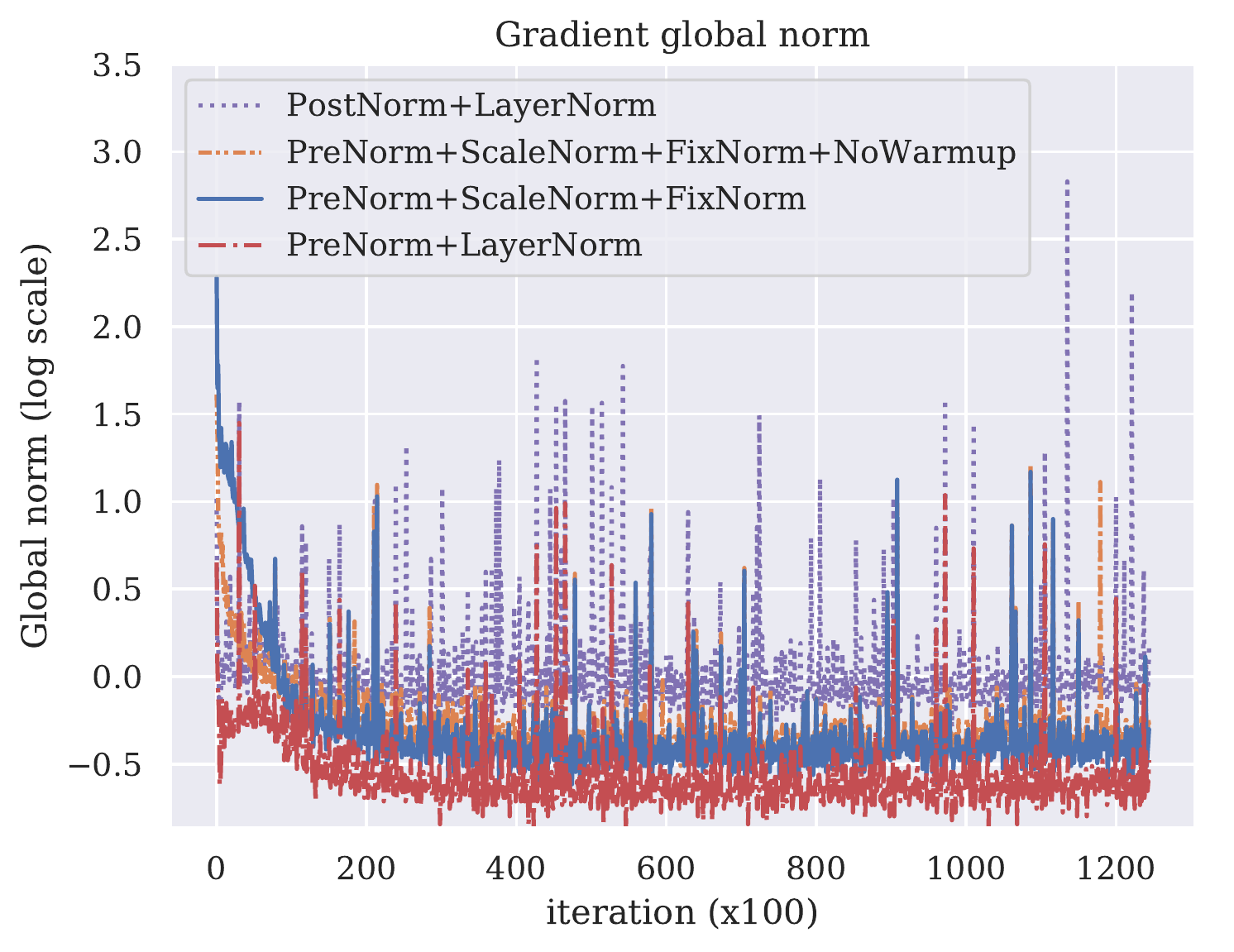}
\caption{The global norm of gradients when using \PostNorm\ or \PreNorm, and with \LNorm, \SCNorm\ and \fixnorm. Best viewed in color.}
\label{fig:gnorm}
\end{figure}

To visualize how \PreNorm\ helps backpropagation, we plot the global gradient norms from our runs in \Cref{fig:gnorm}. \PostNorm\ produces noisy gradients with many sharp spikes, even towards the end of training. On the other hand, \PreNorm\ has fewer noisy gradients with smaller sizes, even without warmup. \LNorm\ has lower global norms than \SCNorm\ + \fixnorm\, but it has more gradient components corresponding to normalization.

%%% <MOVE AROUND FOR OPTIMAL FLOW>
\begin{table*}[ht!]
\small
\begin{minipage}{1.0\linewidth}
    \centering
\begin{tabu}{@{}l|c|c|c|c|c@{}}
\toprule
                 & \multicolumn{1}{c|}{\textbf{\glTOen}}       & \multicolumn{1}{c|}{\textbf{\skTOen}}       & \multicolumn{1}{c|}{\textbf{\enTOvi}}       & \multicolumn{1}{c|}{\textbf{\enTOhe}}       & \multicolumn{1}{c}{\textbf{\arTOen}}       \\ \midrule
\RMSNorm\ + \fixnorm\        & 20.92 & 30.36 & 32.54 & 28.29 & 33.67 \\
\SCNorm\ + \fixnorm\          & 20.91 & 30.25 & 32.79 & 28.44 & 34.15 \\
\SCNormOne\ + \fixnorm\ (learned)          & 21.18 & 30.36 & 32.66 & 28.19 & 34.11 \\
\SCNormOne\ + \fixnorm\ (learned) + \VALBASED\ & 20.36 & 30.45 & 32.83 & 27.97 & 33.98 \\
\SCNormOne\ + \fixnorm\ (learned)  + \VALBASED\ + \lrtwo\ & 21.15 & 30.57 & 31.81 & 25.00 & 28.92 \\
\bottomrule
\end{tabu}
\end{minipage}
\caption{Test BLEU of $\ell_2$-based normalization techniques with different numbers of learned $g$: $O(Ld)$ vs.\ $O(L)$ vs.\ $O(1)$.}
\label{tab:g-ablation}
\end{table*}
%%% </MOVE AROUND FOR OPTIMAL FLOW>

\begin{figure*}[ht!]
\centering
\begin{subfigure}[b]{0.50\textwidth}
    \includegraphics[width=0.9\textwidth]{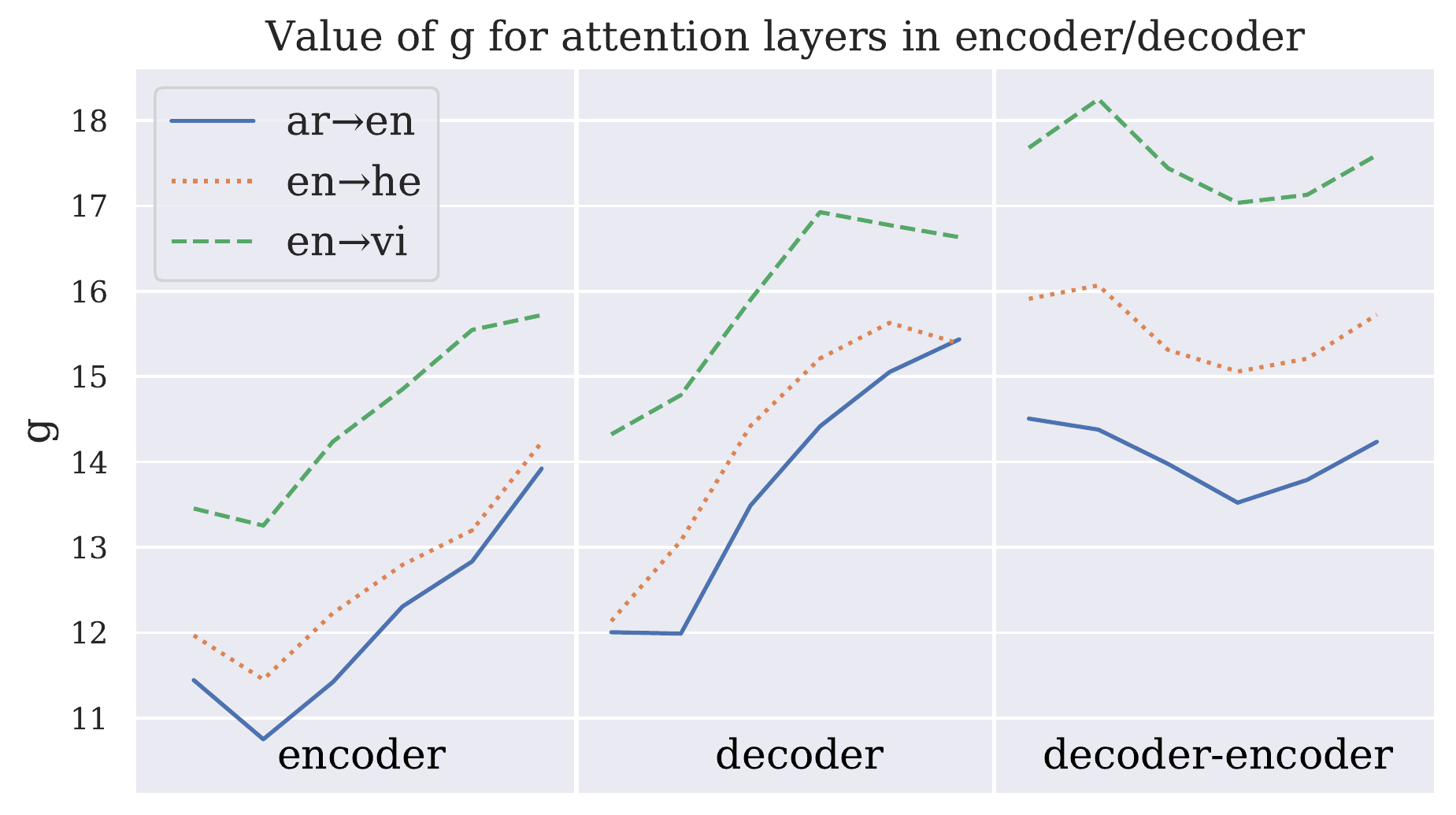}
\end{subfigure}~
\begin{subfigure}[b]{0.50\textwidth}
    \includegraphics[width=0.9\textwidth]{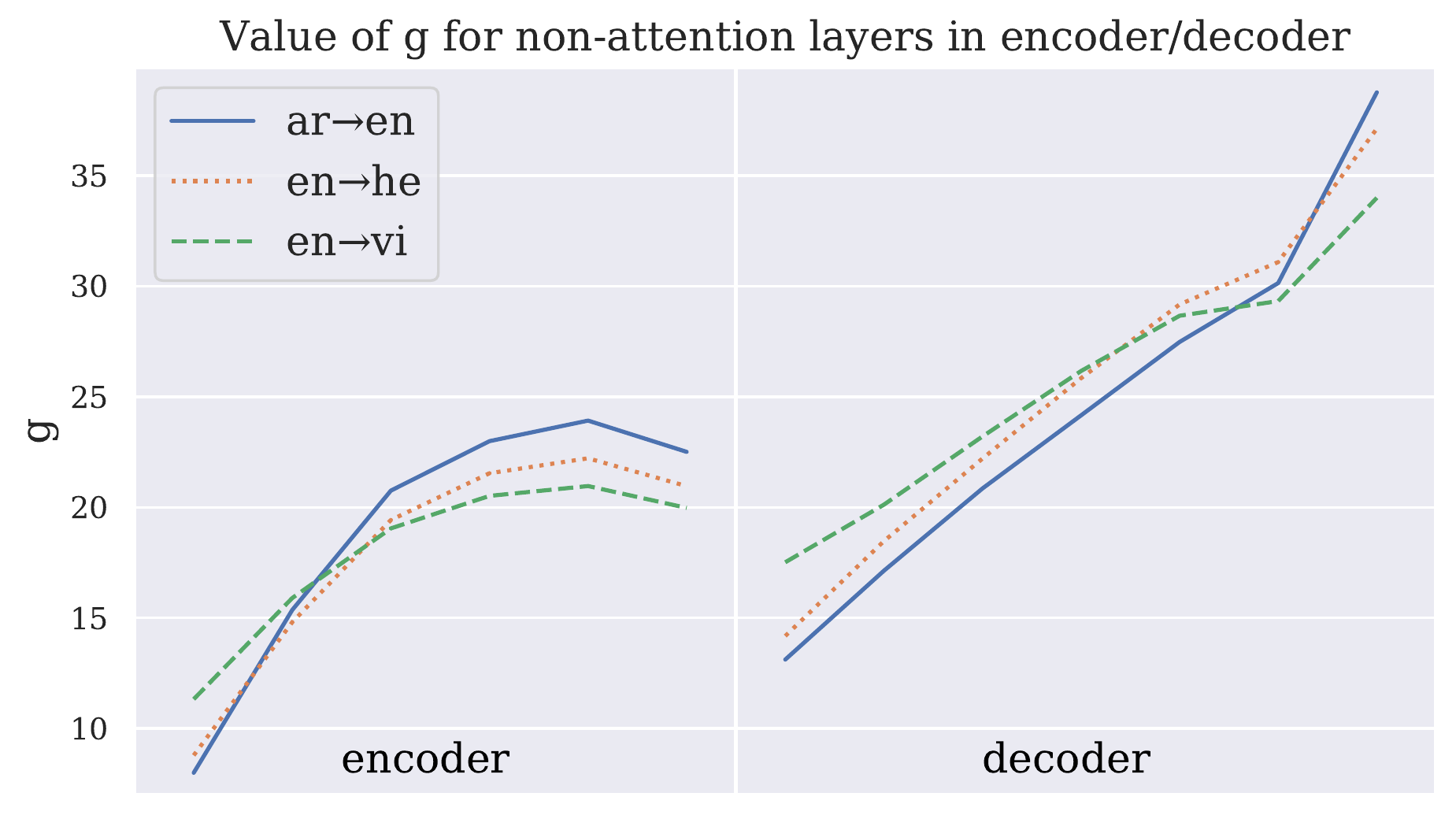}
\end{subfigure}
\caption{Learned $g$ values for \PreNorm\ + \SCNorm\ + \fixnorm\ models, versus depth. \textbf{Left:} Attention sublayers (\keyword{decoder-encoder} denotes decoder sublayers attending on the encoder). \textbf{Right:} Feedforward sublayers and the final linear layer.}
\label{fig:scales}
\end{figure*}

\begin{figure*}[ht!]
\centering
\begin{subfigure}[b]{0.50\textwidth}
    \includegraphics[width=0.9\textwidth]{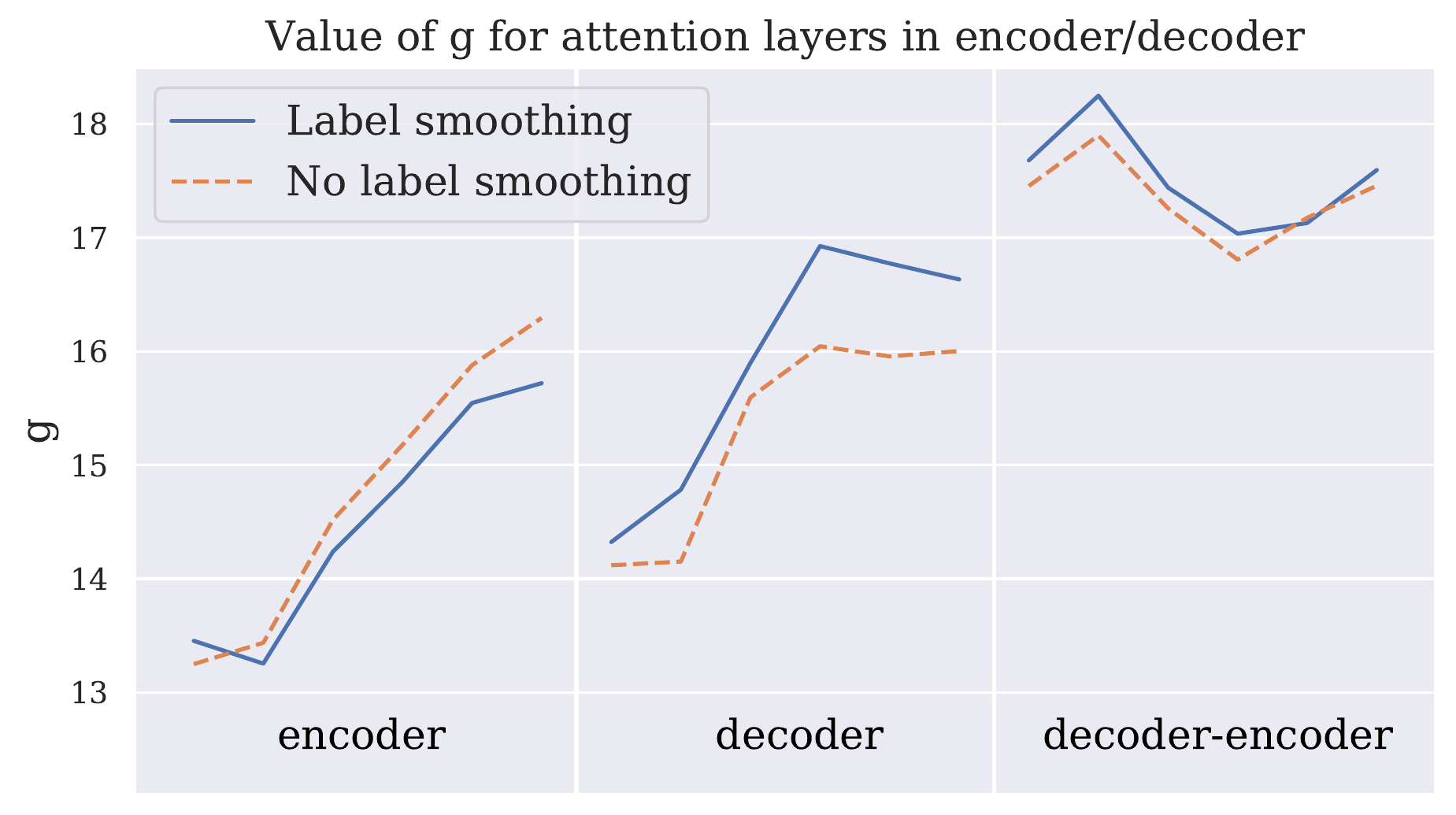}
\end{subfigure}~
\begin{subfigure}[b]{0.50\textwidth}
    \includegraphics[width=0.9\textwidth]{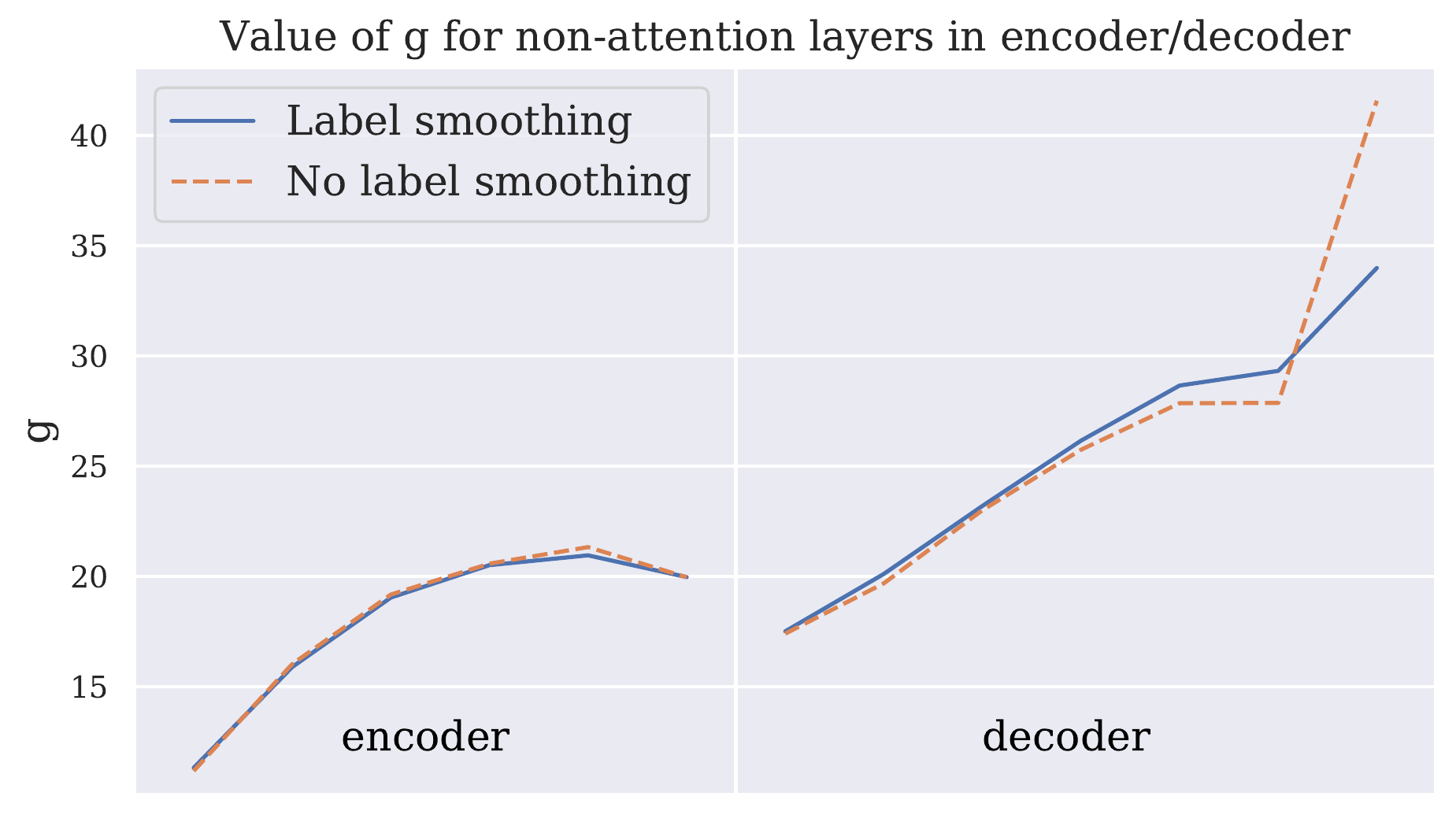}
\end{subfigure}
\caption{Learned $g$ values for our \PreNorm\ + \SCNorm\ + \fixnorm\ \enTOvi\ model (with and without label smoothing), versus depth. \textbf{Left} and \textbf{Right} are the same as in \Cref{fig:scales}.}
\label{fig:scales-ls}
\end{figure*}

\subsection{Activation scaling and the role of $g$}
\label{ssec:g-values}

One motivation for \SCNorm\ was that it expressed a good inductive bias for the global scaling of activations, independent of distributional stability (\Cref{ssec:scaled-cosine}). In contrast, a contemporaneous work \cite{Zhang2019} proposes \keyword{root mean square layer normalization} (\RMSNorm), which still follows layer normalization's motivation but reduces overhead by forgoing additive adjustments, using only a scaling $g_i$ per activation $a_i$. Despite their differing motives, tying the $g_i$ of \RMSNorm\ and dividing by $\sqrt{d}$ retrieves \SCNorm.

Hence we can frame our comparisons in terms of number of learnable parameters. We rerun our \PreNorm\ experiments with \RMSNorm. We also consider fixing $g=\sqrt{d}$ for \SCNorm, where only \fixnorm\ has learnable $g$. \Cref{tab:g-ablation} shows that \SCNorm\ always performs comparably or better than \RMSNorm. Surprisingly, the fixed-$g$ model performs comparably to the one with learnable $g$. However, at higher learning rates (\VALBASED\ with and without \lrtwo), fixed-$g$ models perform much worse on \arTOen, \enTOhe\, and \enTOvi. We conjecture that learning $g$ is required to accommodate layer gradients.

 In \Cref{fig:scales}, we plot the learned $g$ values for pairs with 100k+ examples. For all but the decoder-encoder sublayers, we observe a positive correlation between depth and $g$, giving credence to \SCNorm's inductive bias of global scaling. This trend is clearest in the decoder, where $g$ linearly scales up to the output layer, perhaps in tandem with the discriminativeness of the hidden representations \cite{LiangHL18}. We also note a negative correlation between the number of training examples and the magnitude of $g$ for attention sublayers, which may reflect overfitting.

Finally, to affirm our intuition for interpreting $g$, we plot $g$ values with and without label smoothing (\Cref{fig:scales-ls}). We see a difference in later layers of the decoder; there, removing label smoothing results in lower $g$ values except at the output layer, where $g$ increases sharply. This corresponds to the known overconfidence of translation models' logits, on which label smoothing has a downscaling effect \cite{Muller2019}.

\section{Conclusion}
In this work, we presented three simple, normalization-centric changes to the Transformer model, with a focus on NMT. First, we show that while \PostNorm\ performs better for high-resource NMT in the original base Transformer regime, \PreNorm\ is both more stable and more competent in low-resource settings. Second, we propose replacing \LNorm\ with \SCNorm, a fast and effective \keyword{scaled $\ell_2$ normalization} technique which requires only a single learned parameter. Finally, we reaffirm the effectiveness of fixing the word embedding norm (\fixnorm). Altogether, \PreNorm\ + \fixnorm\ + \SCNorm\ significantly improves NMT on low-resource pairs, with the latter two performing comparably in the high-resource setting, but faster.

In the future, we would like to investigate the relationship between \PostNorm\ and \PreNorm\ when using other optimizers such as \radam\ \cite{radam}, which has been shown to improve Transformer training without warmup. We are also interested in seeing if \fixnorm\ or \SCNorm\ at the final linear layer remains effective when paired with an initialization method such as \fixup\ \cite{fixupinit}, which enables the training of deep neural networks without normalization. One could also explore using other $\ell_p$ norms \cite{Santurkar2018}.

\section*{Acknowledgements}
The authors would like to thank David Chiang and Katrin Kirchhoff for their support of this research.
 
\bibliography{paper}

\begin{thebibliography}{42}
\expandafter\ifx\csname natexlab\endcsname\relax\def\natexlab#1{#1}\fi

\bibitem[{Aharoni et~al.(2019)Aharoni, Johnson, and Firat}]{Aharoni2019}
Roee Aharoni, Melvin Johnson, and Orhan Firat. 2019.
\newblock \href {https://doi.org/10.18653/v1/N19-1388} {{Massively Multilingual
  Neural Machine Translation}}.
\newblock In \emph{NAACL-HLT}, pages 3874--3884.

\bibitem[{Ba et~al.(2015)Ba, Kiros, and Hinton}]{Ba2015}
Jimmy~Lei Ba, Jamie~Ryan Kiros, and Geoffrey~E Hinton. 2015.
\newblock \href {http://arxiv.org/abs/arXiv:1607.06450v1} {{Layer
  Normalization}}.
\newblock \emph{CoRR}, abs/1607.06450.

\bibitem[{Cettolo et~al.(2015)Cettolo, Niehues, Bentivogli, Cattoni, and
  Federico}]{Cettolo2015}
Mauro Cettolo, Jan Niehues, Luisa Bentivogli, Roldano Cattoni, and Marcello
  Federico. 2015.
\newblock \href {http://workshop2015.iwslt.org/downloads/IWSLT_2015_EP_0.pdf}
  {{The IWSLT 2015 Evaluation Campaign}}.
\newblock In \emph{IWSLT}, pages 3--4.

\bibitem[{Chen et~al.(2018)Chen, Firat, Bapna, Johnson, Macherey, Foster,
  Jones, Parmar, Shazeer, Vaswani, Uszkoreit, Kaiser, Schuster, Chen, Wu, and
  Hughes}]{Chen2018}
Mia~Xu Chen, Orhan Firat, Ankur Bapna, Melvin Johnson, Wolfgang Macherey,
  George Foster, Llion Jones, Niki Parmar, Noam Shazeer, Ashish Vaswani, Jakob
  Uszkoreit, Lukasz Kaiser, Mike Schuster, Zhifeng Chen, Yonghui Wu, and
  Macduff Hughes. 2018.
\newblock \href {https://doi.org/10.18653/v1/P18-1008} {{The best of both
  worlds: Combining recent advances in neural machine translation}}.
\newblock In \emph{ACL}, pages 76--86.

\bibitem[{Dong et~al.(2018)Dong, Xu, and Xu}]{Dong2018}
Linhao Dong, Shuang Xu, and Bo~Xu. 2018.
\newblock \href {https://doi.org/10.1109/ICASSP.2018.8462506}
  {{Speech-Transformer: A No-Recurrence Sequence-to-Sequence Model for Speech
  Recognition}}.
\newblock In \emph{ICASSP}, pages 5884--5888.

\bibitem[{Glorot and Bengio(2010)}]{Glorot2010}
Xavier Glorot and Yoshua Bengio. 2010.
\newblock \href {https://doi.org/10.1.1.207.2059} {{Understanding the
  difficulty of training deep feedforward neural networks}}.
\newblock In \emph{AISTATS}, pages 249--256.

\bibitem[{Guo et~al.(2019)Guo, Qiu, Liu, Shao, Xue, and Zhang}]{Guo2019}
Qipeng Guo, Xipeng Qiu, Pengfei Liu, Yunfan Shao, Xiangyang Xue, and Zheng
  Zhang. 2019.
\newblock \href {http://arxiv.org/abs/1902.09113} {{Star-Transformer}}.
\newblock In \emph{NAACL-HLT}, pages 1315--1325.

\bibitem[{He et~al.(2016{\natexlab{a}})He, Zhang, Ren, and Sun}]{He2016}
Kaiming He, Xiangyu Zhang, Shaoqing Ren, and Jian Sun. 2016{\natexlab{a}}.
\newblock \href {https://doi.org/10.1109/CVPR.2016.90} {{Deep residual learning
  for image recognition}}.
\newblock In \emph{CVPR}, pages 770--778.

\bibitem[{He et~al.(2016{\natexlab{b}})He, Zhang, Ren, and
  Sun}]{He2016-identity-mappings}
Kaiming He, Xiangyu Zhang, Shaoqing Ren, and Jian Sun. 2016{\natexlab{b}}.
\newblock \href {http://arxiv.org/abs/1603.05027} {{Identity Mappings in Deep
  Residual Networks}}.
\newblock In \emph{ECCV}, pages 630--645.

\bibitem[{Hieber et~al.(2018)Hieber, Domhan, Denkowski, Vilar, Sokolov,
  Clifton, and Post}]{sockeye}
Felix Hieber, Tobias Domhan, Michael Denkowski, David Vilar, Artem Sokolov, Ann
  Clifton, and Matt Post. 2018.
\newblock \href {https://www.aclweb.org/anthology/W18-1820/} {The {S}ockeye
  neural machine translation toolkit}.
\newblock In \emph{AMTA}, pages 200--207.

\bibitem[{Ioffe and Szegedy(2015)}]{Ioffe2015}
Sergey Ioffe and Christian Szegedy. 2015.
\newblock \href {https://doi.org/10.1007/s13398-014-0173-7.2} {{Batch
  Normalization: Accelerating Deep Network Training by Reducing Internal
  Covariate Shift}}.
\newblock In \emph{ICML}, pages 448--456.

\bibitem[{Kingma and Ba(2015)}]{Kingma2014}
Diederik~P Kingma and Jimmy Ba. 2015.
\newblock \href {http://arxiv.org/abs/1412.6980} {{Adam: A Method for
  Stochastic Optimization}}.
\newblock In \emph{ICLR}.

\bibitem[{Koehn(2004)}]{Koehn2004}
Philipp Koehn. 2004.
\newblock \href {https://doi.org/10.1145/2063576.2063688} {{Statistical
  significance tests for machine translation evaluation}}.
\newblock In \emph{EMNLP}, pages 388--395.

\bibitem[{Kool et~al.(2019)Kool, {Van Hoof}, and Welling}]{Kool2019}
Wouter Kool, Herke {Van Hoof}, and Max Welling. 2019.
\newblock \href {http://arxiv.org/abs/1803.08475v3} {{Attention, Learn to Solve
  Routing Problems!}}
\newblock In \emph{ICLR}.

\bibitem[{Liang et~al.(2018)Liang, Huang, and Lipton}]{LiangHL18}
Davis Liang, Zhiheng Huang, and Zachary~C. Lipton. 2018.
\newblock \href {https://doi.org/10.1109/SLT.2018.8639575} {Learning
  noise-invariant representations for robust speech recognition}.
\newblock In \emph{SLT}, pages 56--63.

\bibitem[{Liu et~al.(2019)Liu, Jiang, He, Chen, Liu, Gao, and Han}]{radam}
Liyuan Liu, Haoming Jiang, Pengcheng He, Weizhu Chen, Xiaodong Liu, Jianfeng
  Gao, and Jiawei Han. 2019.
\newblock \href {http://arxiv.org/abs/1908.03265} {On the variance of the
  adaptive learning rate and beyond}.
\newblock \emph{CoRR}, abs/1908.03265.

\bibitem[{Luo et~al.(2018)Luo, Zhan, Wang, and Yang}]{Luo2017}
Chunjie Luo, Jianfeng Zhan, Lei Wang, and Qiang Yang. 2018.
\newblock \href {http://arxiv.org/abs/1702.05870} {{Cosine Normalization: Using
  Cosine Similarity Instead of Dot Product in Neural Networks}}.
\newblock In \emph{ICANN}, pages 382--391.

\bibitem[{M{\"{u}}ller et~al.(2019)M{\"{u}}ller, Kornblith, and
  Hinton}]{Muller2019}
Rafael M{\"{u}}ller, Simon Kornblith, and Geoffrey Hinton. 2019.
\newblock \href {http://arxiv.org/abs/1906.02629} {{When Does Label Smoothing
  Help?}}
\newblock In \emph{NeurIPS}.

\bibitem[{Neubig and Hu(2018)}]{Neubig2019}
Graham Neubig and Junjie Hu. 2018.
\newblock \href {https://doi.org/10.18653/v1/d18-1103} {{Rapid Adaptation of
  Neural Machine Translation to New Languages}}.
\newblock In \emph{EMNLP}, pages 875--880.

\bibitem[{Nguyen and Chiang(2018)}]{Nguyen2018-improving-lexical-choice}
Toan Nguyen and David Chiang. 2018.
\newblock \href {https://doi.org/10.18653/v1/n18-1031} {{Improving Lexical
  Choice in Neural Machine Translation}}.
\newblock In \emph{NAACL-HLT}, pages 334--343.

\bibitem[{Ott et~al.(2019)Ott, Edunov, Baevski, Fan, Gross, Ng, Grangier, and
  Auli}]{fairseq}
Myle Ott, Sergey Edunov, Alexei Baevski, Angela Fan, Sam Gross, Nathan Ng,
  David Grangier, and Michael Auli. 2019.
\newblock \href {https://doi.org/10.18653/v1/n19-4009} {{fairseq: A Fast,
  Extensible Toolkit for Sequence Modeling}}.
\newblock In \emph{NAACL-HLT (Demonstrations)}, pages 48--53.

\bibitem[{Ott et~al.(2018)Ott, Edunov, Grangier, and Auli}]{Ott2018}
Myle Ott, Sergey Edunov, David Grangier, and Michael Auli. 2018.
\newblock \href {https://www.aclweb.org/anthology/W18-6301/} {{Scaling Neural
  Machine Translation}}.
\newblock In \emph{WMT}, pages 1--9.

\bibitem[{Papineni et~al.(2002)Papineni, Roukos, Ward, and
  Zhu}]{papineni-etal-2002-bleu}
Kishore Papineni, Salim Roukos, Todd Ward, and Wei-Jing Zhu. 2002.
\newblock \href {https://doi.org/10.3115/1073083.1073135} {{B}leu: a method for
  automatic evaluation of machine translation}.
\newblock In \emph{ACL}, pages 311--318.

\bibitem[{Pascanu et~al.(2013)Pascanu, Mikolov, and Bengio}]{Pascanu2013}
Razvan Pascanu, Tomas Mikolov, and Yoshua Bengio. 2013.
\newblock \href {https://doi.org/10.1109/72.279181} {{On the difficulty of
  training recurrent neural networks}}.
\newblock In \emph{ICML}, pages 1310--1318.

\bibitem[{Pereyra et~al.(2017)Pereyra, Tucker, Chorowski, Kaiser, and
  Hinton}]{label_smoothing_2}
Gabriel Pereyra, George Tucker, Jan Chorowski, Lukasz Kaiser, and Geoffrey~E.
  Hinton. 2017.
\newblock \href {https://openreview.net/forum?id=HyhbYrGYe} {Regularizing
  neural networks by penalizing confident output distributions}.
\newblock In \emph{ICLR (Workshop)}.

\bibitem[{Popel and Bojar(2018)}]{Popel2018}
Martin Popel and Ondřej Bojar. 2018.
\newblock \href {https://doi.org/10.2478/pralin-2018-0002} {{Training Tips for
  the Transformer Model}}.
\newblock \emph{Prague Bull. Math. Linguistics}, 110(1):43--70.

\bibitem[{Press and Wolf(2017)}]{tiedembed}
Ofir Press and Lior Wolf. 2017.
\newblock \href {https://www.aclweb.org/anthology/E17-2025/} {{Using the Output
  Embedding to Improve Language Models}}.
\newblock In \emph{EACL}, pages 157--163.

\bibitem[{Qi et~al.(2018)Qi, Sachan, Felix, Padmanabhan, and
  Neubig}]{Qi2018-word-embeddings-nmt}
Ye~Qi, Devendra Sachan, Matthieu Felix, Sarguna Padmanabhan, and Graham Neubig.
  2018.
\newblock \href {https://doi.org/10.18653/v1/n18-2084} {{When and Why Are
  Pre-Trained Word Embeddings Useful for Neural Machine Translation?}}
\newblock In \emph{NAACL-HLT}, pages 529--535.

\bibitem[{Salazar et~al.(2019)Salazar, Kirchhoff, and Huang}]{Salazar2019}
Julian Salazar, Katrin Kirchhoff, and Zhiheng Huang. 2019.
\newblock \href {https://doi.org/10.1109/ICASSP.2019.8682539} {{Self-attention
  Networks for Connectionist Temporal Classification in Speech Recognition}}.
\newblock In \emph{ICASSP}, pages 7115--7119.

\bibitem[{Santurkar et~al.(2018)Santurkar, Tsipras, Ilyas, and
  Madry}]{Santurkar2018}
Shibani Santurkar, Dimitris Tsipras, Andrew Ilyas, and Aleksander Madry. 2018.
\newblock \href
  {http://papers.nips.cc/paper/7515-how-does-batch-normalization-help-optimization}
  {{How does batch normalization help optimization?}}
\newblock In \emph{NeurIPS}, pages 2483--2493.

\bibitem[{Sennrich et~al.(2016{\natexlab{a}})Sennrich, Haddow, and
  Birch}]{Sennrich2016dropout}
Rico Sennrich, Barry Haddow, and Alexandra Birch. 2016{\natexlab{a}}.
\newblock \href {http://arxiv.org/abs/1606.02891} {{Edinburgh Neural Machine
  Translation Systems for WMT 16}}.
\newblock In \emph{WMT}, pages 371--376.

\bibitem[{Sennrich et~al.(2016{\natexlab{b}})Sennrich, Haddow, and
  Birch}]{Sennrich2016}
Rico Sennrich, Barry Haddow, and Alexandra Birch. 2016{\natexlab{b}}.
\newblock \href {https://www.aclweb.org/anthology/P16-1162/} {{Neural machine
  translation of rare words with subword units}}.
\newblock In \emph{ACL}.

\bibitem[{Shazeer and Stern(2018)}]{Shazeer2018}
Noam Shazeer and Mitchell Stern. 2018.
\newblock \href {http://arxiv.org/abs/1804.04235} {{Adafactor: Adaptive
  Learning Rates with Sublinear Memory Cost}}.
\newblock In \emph{ICML}, pages 4603--4611.

\bibitem[{Sukhbaatar et~al.(2019)Sukhbaatar, Grave, Lample, Jegou, and
  Joulin}]{Sukhbaatar2019}
Sainbayar Sukhbaatar, Edouard Grave, Guillaume Lample, Herve Jegou, and Armand
  Joulin. 2019.
\newblock \href {http://arxiv.org/abs/1907.01470} {{Augmenting Self-attention
  with Persistent Memory}}.
\newblock \emph{CoRR}, abs/1907.01470.

\bibitem[{Szegedy et~al.(2016)Szegedy, Vanhoucke, Ioffe, Shlens, and
  Wojna}]{label_smoothing_1}
Christian Szegedy, Vincent Vanhoucke, Sergey Ioffe, Jon Shlens, and Zbigniew
  Wojna. 2016.
\newblock \href {https://doi.org/10.1109/CVPR.2016.308} {{Rethinking the
  Inception Architecture for Computer Vision}}.
\newblock In \emph{CVPR}, pages 2818--2826.

\bibitem[{Vaswani et~al.(2018)Vaswani, Bengio, Brevdo, Chollet, Gomez, Gouws,
  Jones, Kaiser, Kalchbrenner, Parmar, Sepassi, Shazeer, and
  Uszkoreit}]{tensor2tensor}
Ashish Vaswani, Samy Bengio, Eugene Brevdo, Francois Chollet, Aidan~N Gomez,
  Stephan Gouws, Llion Jones, {\L}ukasz Kaiser, Nal Kalchbrenner, Niki Parmar,
  Ryan Sepassi, Noam Shazeer, and Jakob Uszkoreit. 2018.
\newblock \href {http://arxiv.org/abs/1803.07416} {{Tensor2Tensor for Neural
  Machine Translation}}.
\newblock In \emph{AMTA}, pages 193--199.

\bibitem[{Vaswani et~al.(2017)Vaswani, Shazeer, Parmar, Uszkoreit, Jones,
  Gomez, Kaiser, and Polosukhin}]{NIPS2017_7181}
Ashish Vaswani, Noam Shazeer, Niki Parmar, Jakob Uszkoreit, Llion Jones,
  Aidan~N Gomez, {\L}ukasz Kaiser, and Illia Polosukhin. 2017.
\newblock \href {http://papers.nips.cc/paper/7181-attention-is-all-you-need}
  {{Attention is All you Need}}.
\newblock In \emph{NeurIPS}, pages 5998--6008.

\bibitem[{Veit et~al.(2016)Veit, Wilber, and Belongie}]{resnetensemble}
Andreas Veit, Michael Wilber, and Serge Belongie. 2016.
\newblock \href
  {http://papers.nips.cc/paper/6556-residual-networks-behave-like-ensembles-of-relatively-shallow-networks}
  {{Residual networks behave like ensembles of relatively shallow networks}}.
\newblock \emph{NeurIPS}, pages 550--558.

\bibitem[{Wang et~al.(2019)Wang, Li, Xiao, Zhu, Li, Wong, and
  Chao}]{Wang2019-learning-deep-transformers}
Qiang Wang, Bei Li, Tong Xiao, Jingbo Zhu, Changliang Li, Derek~F Wong, and
  Lidia~S Chao. 2019.
\newblock \href {https://www.aclweb.org/anthology/P19-1176/} {{Learning Deep
  Transformer Models for Machine Translation}}.
\newblock In \emph{ACL}, pages 1810--1822.

\bibitem[{Wang et~al.(2018)Wang, Pham, Dai, and Neubig}]{Wang2018-switchout}
Xinyi Wang, Hieu Pham, Zihang Dai, and Graham Neubig. 2018.
\newblock \href {https://www.aclweb.org/anthology/D18-1100/} {Switchout: an
  efficient data augmentation algorithm for neural machine translation}.
\newblock In \emph{EMNLP}, pages 856--861.

\bibitem[{Zhang and Sennrich(2019)}]{Zhang2019}
Biao Zhang and Rico Sennrich. 2019.
\newblock \href {https://openreview.net/pdf?id=SygkZ3MTJE} {{Root Mean Square
  Layer Normalization}}.
\newblock \emph{NeurIPS}.

\bibitem[{Zhang et~al.(2019)Zhang, Dauphin, and Ma}]{fixupinit}
Hongyi Zhang, Yann~N. Dauphin, and Tengyu Ma. 2019.
\newblock \href {https://openreview.net/forum?id=H1gsz30cKX} {Fixup
  initialization: Residual learning without normalization}.
\newblock In \emph{ICLR}.

\end{thebibliography}
\bibliographystyle{acl_natbib}

\appendix

\section{Training details}
\label{appendix:setup}

\paragraph{Data and preprocessing.} The pairs are English (en) to {Hebrew (he), Vietnamese (vi)}, and {Galician (gl), Slovak (sk), Arabic (ar)} to English (en). Because the data is already preprocessed, we only apply BPE \cite{Sennrich2016} with \texttt{fastBPE}\footnote{\url{https://github.com/glample/fastBPE}}. Depending on the data size, we use different numbers of BPE operations.

We wanted to compare with the latest low-resource works of \cite{Neubig2019, Aharoni2019} on the TED Talks corpus \cite{Qi2018-word-embeddings-nmt}. In particular, \citet{Aharoni2019} identified 4 very low-resource pairs ($<$70k); we took the two (\glTOen, \skTOen) that were not extremely low ($\le$6k). They then identified 4 low-resource pairs with 100k-300k examples; we took the top two (\arTOen, \enTOhe). To introduce a second English-source pair and to showcase on a well-understood task, we used the \enTOvi\ pair from \IWSLT\ with an in-between number of examples (133k). In this way, we have examples of different resource levels, language families, writing directions, and English-source versus -target.

\paragraph{Model configuration.} We set the hidden dimension of the feedforward sublayer to 2048 and the rest to 512, matching \citet{NIPS2017_7181}. We use the same dropout rate for output of sublayers, ReLU, and attention weights. Additionally, we also do word dropout \cite{Sennrich2016dropout} with probability 0.1. However, instead of zeroing the word embeddings, we randomly replace tokens with \texttt{UNK}. For all experiments, we use label smoothing of 0.1 \cite{label_smoothing_1, label_smoothing_2}. The source and target's input and output embeddings are shared \cite{tiedembed}, but we mask out words that are not in the target's vocabulary at the final output layer before softmax, by setting their logits to $-\infty$.

\paragraph{Training.} We use a batch size of 4096 and optimize using Adam \cite{Kingma2014} with the default parameters $\beta_1=0.9$, $\beta_2=0.999$, $\epsilon=10^{-8}$. Gradients are clipped when global norm exceeds 1.0 \cite{Pascanu2013}. An epoch is a predefined number of iterations for each pair. We stop training when a maximum number of epochs has been met or the learning rate becomes too small ($10^{-6}$). We also do early stopping when the development BLEU has not improved for 20 evaluations. For \glTOen, this number is 50. When doing validation-based decay, we use $\alpha_{decay} = 0.8$ and $patience = 3$. For complete data and model statistics, please refer to \Cref{tab:stats}. The best checkpoint is selected based on the development BLEU score during training.

\paragraph{Evaluation.} We report tokenized BLEU \cite{papineni-etal-2002-bleu} with \texttt{multi-bleu.perl} to be comparable with previous works. We also measure statistical significance using bootstrap resampling \cite{Koehn2004}. For \WMT\ English-German, note that one needs to put compounds in ATAT format\footnote{\url{https://github.com/tensorflow/tensor2tensor/blob/master/tensor2tensor/utils/get_ende_bleu.sh}} before calculating BLEU score to be comparable with previous works.\\

\section{Further analysis}
\label{ssec:generalization}

%%% <MOVE AROUND FOR OPTIMAL FLOW>
\begin{table}[h!]
\small
\begin{minipage}{1.0\linewidth}
    \centering
\begin{tabu}{@{}c|cc|cc@{}}
\toprule
      & \multicolumn{2}{c|}{\LNorm}        & \multicolumn{2}{c}{\SCNorm}       \\
      & \textbf{train} & \textbf{test} & \textbf{train} & \textbf{test} \\ \midrule
\textbf{\glTOen} & 11.792 & 54.300  & 10.151 & 45.770 \\
\textbf{\skTOen} & 14.078 & 20.460 & 14.004 & 19.080 \\
\textbf{\enTOvi} & 15.961 & 17.950 & 16.719 & 17.100  \\
\textbf{\enTOhe} & 15.562 & 14.950 & 15.906 & 15.080 \\
\textbf{\arTOen} & 14.372 & 13.450 & 14.165 & 13.290 \\
\bottomrule
\end{tabu}
\end{minipage}
\caption{Label-smoothed train/test perplexities when using \LNorm\ and \SCNorm. }
\label{tab:train-test-ppl}
\end{table}
%%% </MOVE AROUND FOR OPTIMAL FLOW>

We ask if improvements from \SCNorm\ on our low-resource tasks are due to improved regularization (a smaller generalization gap) or improved overall performance. We record smoothed train and test perplexities of our \PreNorm\ models in \Cref{tab:train-test-ppl}. We see suggestive results but no conclusive trends. For \arTOen, \glTOen, and \skTOen, train and test drop slightly, with test more so than train. For \enTOvi, train perplexity increases and test perplexity decreases an equivalent amount. For \enTOhe, our smallest change between \SCNorm\ and \LNorm, train perplexity negligibly increased and test perplexity remains the same.

\section{Listings}

See the following page.

\onecolumn

\paragraph{\SCNorm.}
\small
\begin{lstlisting}[language=Python]
  class ScaleNorm(nn.Module):
    """ScaleNorm"""
    def __init__(self, scale, eps=1e-5):
        super(ScaleNorm, self).__init__()
        self.scale = Parameter(torch.tensor(scale))
        self.eps = eps

    def forward(self, x):
        norm = self.scale / torch.norm(x, dim=-1, keepdim=True).clamp(min=self.eps)
        return x * norm
\end{lstlisting}
\normalsize

\paragraph{\FairSeq.}
\label{listing-fairseq}
We follow \FairSeq's tutorial\footnote{\url{https://github.com/pytorch/fairseq/blob/master/examples/scaling_nmt/README.md}} and train a \PostNorm\ Transformer base model using the following configuration:

\small
\begin{lstlisting}[language=bash]
  fairseq-train \
  data-bin/wmt16_en_de_bpe32k/ \
  --arch transformer_wmt_en_de \
  --share-all-embeddings \
  --optimizer adam \
  --adam-betas '(0.9, 0.98)' \
  --clip-norm 1.0 \
  --lr 0.001 \
  --lr-scheduler inverse_sqrt \
  --warmup-updates 4000 \
  --warmup-init-lr 1e-07 \
  --dropout 0.1 \
  --weight-decay 0.0 \
  --criterion label_smoothed_cross_entropy \
  --label-smoothing 0.1 \
  --max-tokens 8192 \
  --update-freq 10 \
  --attention-dropout 0.1 \
  --activation-dropout 0.1 \
  --max-epoch 40
\end{lstlisting}
\normalsize

For \PreNorm, simply include the flags:
\small
\begin{lstlisting}[language=bash]
	--encoder-normalize-before --decoder-normalize-before
\end{lstlisting}
\normalsize

For \SCNorm, we replace all {\LNorm}s in {\texttt{fairseq/models/transformer.py}} and {\texttt{fairseq/modules/transformer\_layer.py}} with \SCNorm\ (implemented above). For \fixnorm, we change the word embedding initialization to uniform with range $[-0.01, 0.01]$ and normalize with {\lstinline{torch.nn.functional.normalize}}.  

We note that \FairSeq\ uses Xavier uniform initialization, which is big compared to our \SmallInit\ (\Cref{experiment_weight_init}). We conjecture that \FairSeq\ training remains stable thanks to its large batch size, which gives more stable gradients.

\end{document}